\pdfoutput=1

\documentclass[11pt]{article}

\usepackage[review]{acl}

\usepackage{times}
\usepackage{latexsym}
\usepackage{amsfonts}
\usepackage[T1]{fontenc}

\usepackage[utf8]{inputenc}

\usepackage{microtype}

\usepackage{graphicx}
\usepackage{amsmath}
\usepackage{booktabs}
\usepackage[ruled,vlined ]{algorithm2e}
\usepackage[noend]{algpseudocode}
\usepackage{amsmath}
\usepackage{enumitem}
\usepackage{bm}
\usepackage{subcaption}
\usepackage{wrapfig}

\usepackage{etoolbox}
\title{Improving Gender Fairness of Pre-Trained Language Models without Catastrophic Forgetting}

\newcommand*{\affaddr}[1]{#1} 
\newcommand*{\affmark}[1][*]{\textsuperscript{#1}}

\author{%
Zahra Fatemi\affmark[1]\thanks{\ \ Work done during internship with Salesforce Research} , Chen Xing\affmark[2], Wenhao Liu\affmark[2], Caiming Xiong\affmark[2]\\
\affaddr{\affmark[1]Department of Computer Science, University of Illinois Chicago}\\
\affaddr{\affmark[2]Salesforce Research}\\
 \texttt{zfatem2@uic.edu}\\
\texttt{\{cxing,wenhao.liu,cxiong\}@salesforce.com}
}

\begin{document}
{\makeatletter\acl@finalcopytrue
  \maketitle
}
\nolinenumbers
\begin{abstract}
Existing studies addressing gender bias of pre-trained language models, usually build a small gender-neutral data set and conduct a second phase pre-training on the  model with such data. However, given the limited size and concentrated focus of the gender-neutral data, catastrophic forgetting would occur during second-phase pre-training. Forgetting information in the original training data may damage the model’s downstream performance by a large margin. In this work, we empirically show that catastrophic forgetting occurs in such methods by evaluating them with general NLP tasks in GLUE. Then, we propose a new method, GEnder Equality Prompt (GEEP), to improve gender fairness of pre-trained models with less forgetting. GEEP freezes the pre-trained model and learns gender-related prompts with gender-neutral data.
Empirical results show that GEEP not only achieves SOTA performances on gender fairness tasks, but also forgets less and performs better on GLUE by a large margin.

\end{abstract}
\section{Introduction}
Pre-trained language models, e.g., BERT \citep{devlin-naacl19} and RoBERTa \citep{liu-arxiv19}, have shown competitive performance in a wide variety of NLP downstream applications.
However, such models are often prone to exhibit gender bias \citep{de-acl21,zhao-acl19,webster-arxiv20}, due to their large scale unsupervised training data from the web \citep{liu-arxiv19,brown2020language}. Gender bias refers to
unbalanced model behaviors with respect to a specific gender \citep{cheng-iclr21}.
Among various gender-biased behaviours of pre-trained models, bias on professions is the most prominent and well-studied \citep{de-acl21,vig-neurips20,qian-acl19,zhao-acl19}.
For example, in coreference resolution tasks, a pre-trained model would predict female pronoun and names for professions like ``nurse'' and ``housekeeper'', while predict male pronouns for ``computer programmer'' or ``doctor'' \citep{kurita-gbnlp19}. The pre-trained models also wouldn't prefer gender-neutral pronouns actively, which is unfair to other gender identities beyond males/females \citep{deutsch2015electronic}. 

Given the large model size and tremendous time complexity for language model pre-training, training a gender-neutral model from scratch with manually filtered data seems impossible for most organizations.
Due to this limitation, existing studies usually build a relatively small gender-neutral data set (for example building a data set that have more balanced gender pronouns for profession names), and conduct second phase pre-training on the pre-trained model with such data \citep{webster-arxiv20, de-acl21}. However, given the limited size of the gender-neutral data and its potential distributional mismatch with the original pre-training data, \textit{catastrophic forgetting} can occur during second-phase pre-training of such methods. Catastrophic forgetting \citep{kirkpatrick-nas17} is a long-standing problem which illustrates the tendency of a neural network to forget previously learned information upon learning new information. When it comes to further training a pre-trained model, using the small gender-neutral data to update the entire massive model could make the model forget the diverse information from the original pre-training data, which damages the model's downstream performance by a large margin.

In this paper, we first empirically verify that further updating a pre-trained model (such as RoBERTa \citep{liu-arxiv19}) with manually-built gender-neutral data can cause catastrophic forgetting. We follow existing work and build our profession-related gender-neutral data set by filtering out Wikipedia sentences mentioning professions and swapping their gender related pronouns.
We find that although our gender-neutral data is from Wikipedia which is part of RoBERTa's pre-training data, the model's performance on downstream tasks in GLUE \citep{wang-emnlp18} still drops with a considerable margin after second-phase pre-training, due to the smaller size and more concentrated focus of the gender-neutral data.

Therefore, we propose a new method, GEnder Equality Prompt (GEEP), to alleviate gender bias of pre-trained models without catastrophic forgetting. 
Specifically, inspired by recent prompt-tuning methods \citep{lester-arxiv21} for fine-tuning large pre-trained models, GEEP freezes the entire model, adds and updates new word embeddings of professions as gender equality prompts, instead of updating all model parameters at second-phase pre-training as previous methods.
Since all the pre-trained parameters are frozen during further training, diverse information from the original training data preserved in the pre-trained parameters is not erased. Therefore forgetting can be alleviated to large extent. 
Moreover,
since the embeddings of professions are re-initialized when debiasing training starts, gender bias from previous data that is embedded in such representations is already removed before second-phase pre-training. 
Therefore, GEEP also improves gender fairness of the model more effectively with much fewer iterations. 
Empirical results show that GEEP not only achieves state-of-the-art performances with fewer iterations on various gender fairness tasks such as pronoun coreference resolution, but also forgets less and achieves better results on GLUE tasks. 

\begin{figure*}
    \centering
        \includegraphics[width=0.5\textwidth]{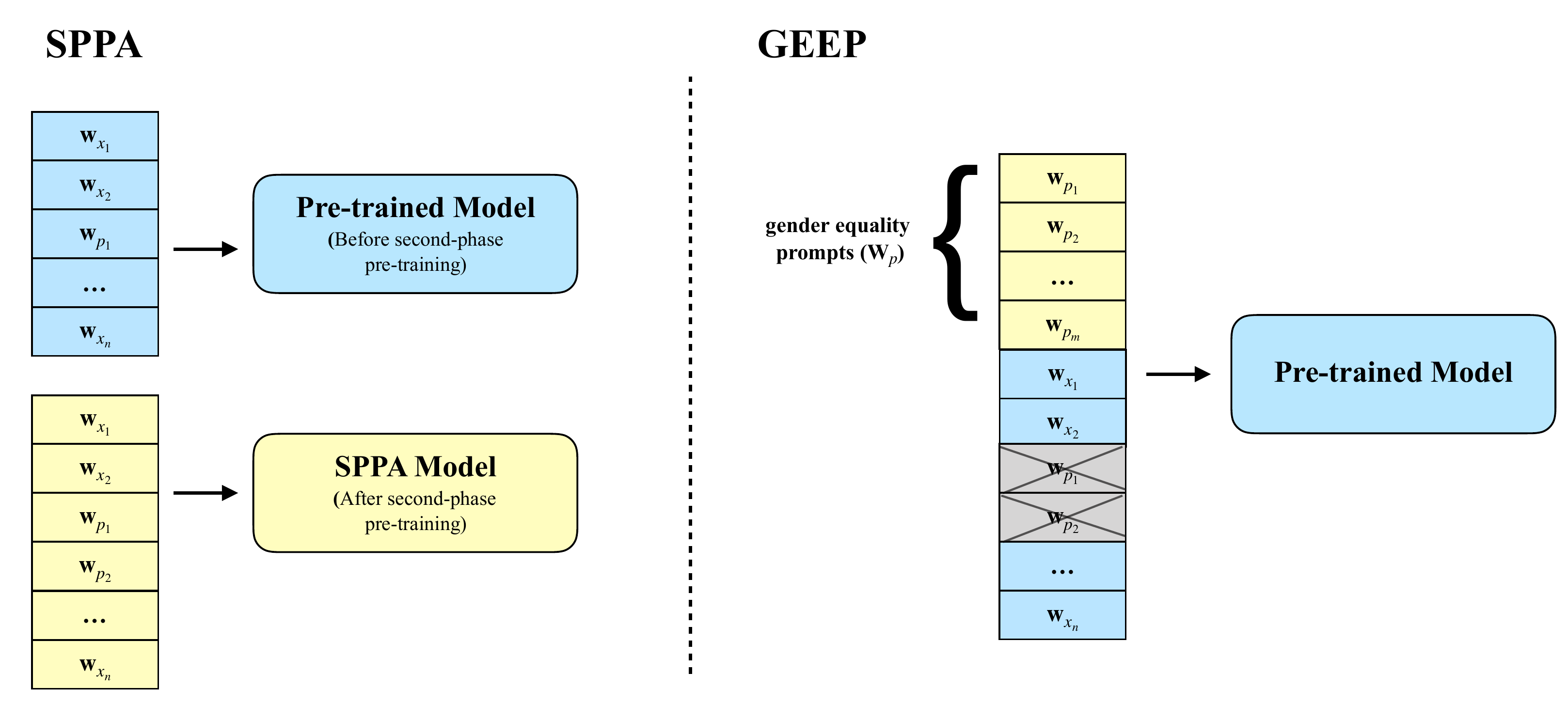}
        \vspace{-3mm}
        \caption{
        Difference between SPPA and GEEP methods. Blue boxes represent the parameters of the pre-trained model before any further training and yellow boxes show updated parameters during second-phase pre-training (SPPA). SPPA requires updating all the pre-trained model's parameters. In contrast, GEEP only adds and updates new embeddings of biased professions such as $\mathbf{w}_{p_i}$. Gray boxes are the original embeddings of professions which are not updated/used in second phase pre-training or the training/inference after that.
       }
    \label{fig:geep_diagram}
     \vspace{-3mm}
\end{figure*}

\section{Related Work}

Compared with the existing work focusing on quantifying and alleviating gender bias \citep{bolukbasi-neurips16,caliskan-aaas17,zhao-emnlp18,gonen-acl19,sun-acl19,garg-nas18,zhao-acl18, bolukbasi-neurips16, zhao-emnlp18} in standard word embedding models, such as
word2vec \citep{mikolov-neurips13} and GloVe \citep{pennington-emnlp14}, gender bias in large pre-trained language models seems less studied.
Recent work on gender fairness of pre-trained language models, such as ELMo \citep{peters-naacl18} and BERT \citep{devlin-naacl19}, mostly focus on showing and measuring the gender bias embedded in such models \citep{zhao-acl19,tan-neurips20}.  
These studies propose metrics to quantify gender bias in pre-trained language models \citep{de-acl21,tan-neurips20,webster-acl18,kurita-gbnlp19}. In our work, we employ such methods to evaluate GEEP and baseline methods on improving gender fairness.
Existing works focusing on mitigating gender bias of pre-trained models usually collect and build gender-neutral data on their own and conduct a second phase pre-training on the released pre-trained model \citep{webster-arxiv20, de-acl21, cheng-iclr21}.
In this work, we demonstrate empirically that the performance of the debiased model on general downstream tasks such as GLUE, still drops by a considerable margin after such second-phase pre-training. Then, given this phenomenon, we propose GEEP to alleviate gender bias in pre-trained models without forgetting.

\section{Improving Gender Fairness without Forgetting}
In this section, we first describe the gender-neutral collection method we adopt from existing methods and the forgetting issue in such methods. Then we describe the proposed method GEnder Equality Prompt (GEEP).

\subsection{Profession-Related Gender-Neutral Data Collection}

We follow existing work to build a profession-related gender neutral data set since profession-related gender bias is a relatively well-studied aspect of gender bias.
To construct profession-related data with equal numbers of references to male and female genders, we adopt the data filtering method by \cite{zhao-acl18} on the English Wikipedia corpus.
Specifically, we filter Wikipedia for sentences containing at least one profession that is supposed to be gender-neutral but generally viewed with gender bias, e.g., nurse, defined by \cite{bolukbasi-neurips16}. 
For each of these sentences, we swap the gendered terms with their opposite genders (such as ``Man'' \textrightarrow  ``Woman'', ``he''\textrightarrow ``she'', and vice-versa). 
We also provide an analysis of the processed data in Appendix~\ref{appx:grammar}.
Our dataset includes both the original profession-related sentences and their gender-swapped counterparts.
We get $6.1$GB of profession-related gender-neutral text data. Compared with the original pre-training data of RoBERTa (160GB in text size from various sources), the gender-neutral data we have is smaller and less diverse. 

After the gender-neutral data set is built, a common approach to mitigate gender bias in pre-trained language models is to conduct second-phase pre-training to update all model parameters with this data set. We refer to such methods as \textit{SPPA} (Second-Phase Pre-training for All parameters). 
In Section \ref{experiments}, we empirically show that SPPA methods lead to forgetting and the model's performance on NLP benchmark GLUE drops by a large margin.

\subsection{Gender Equality Prompt Approach}

To alleviate forgetting while mitigating gender bias in pre-trained language models, we propose GEnder Equality Prompt (GEEP). 
In GEEP, instead of updating all model parameters during second-phase pre-training, we freeze all of the pre-trained model parameters and add new trainable embeddings for profession names as gender equality prompts.
Since all previous pre-trained parameters are frozen, diverse information from original massive pre-training data that are memorized by the pre-trained parameters wouldn't be erased. 
Therefore, the forgetting of information from the original training data can be alleviated to the fullest extent. 

Let $\mathbf{X}=\{x_1,x_2,...,x_n \}$ denote the original vocabulary of the pre-trained model and $\mathbf{W}_{x} \in \mathbb{R}^{n \times d}$ be the original pre-trained token embedding matrix of the model with dimension of $d$. Given a set of $m$ profession names, $\{p_1,p_2,...,p_m\}$, we build an embedding matrix $\mathbf{W}_{p} \in \mathbb{R}^{m \times d}$ where the embedding of each token is initialized randomly.  To obtain an integrated word embedding matrix, we concatenate $\mathbf{W}_x$ and $\mathbf{W_p}$ as $\mathbf{W}_{\text{emb}}=\text{Concat}(\mathbf{W}_x,\mathbf{W}_p)$. We note that we concatenate them along the dimension of words/tokens instead of in the embedding space. After concatenation, the model's representation size (hidden) remain unchanged.
During both second-phase pre-training and the training/inference after that, once a profession occurs, we only update/use its new embedding in $\mathbf{W}_p$. We show the comparison between GEEP and other second-phase pre-training methods in Figure \ref{fig:geep_diagram}.
Given all the pre-trained model's frozen parameters $\mathbf{W}_{\text{whole}}$ that contains $\mathbf{W}_x$, the objective function of second-phase pre-training of GEEP is,
\begin{align}
\small
\mathcal{L}&(\mathbf{x}_{\text{masked}}| \mathbf{x}_{\text{context}}, \mathbf{W}_{\text{whole}})\\
&= \frac{1}{N_{\text{mask}}} (\sum_{t=1}^{N_{\text{mask}}} -\log p_{\theta}(x_t|\mathbf{x}_{\text{context}}, \mathbf{W}_{\text{whole}})).
\label{eq:bert-loss}
\vspace{-5mm}
\end{align}

$N_{\text{mask}}$ is the number of masked positions in the input sequence $\mathbf{x}$. With such an objective, $\mathbf{W}_p$ is updated with gender-neutral data. Moreover, since the embeddings of professions are re-initialized when debiasing training starts in GEEP, gender bias from previous data that is embedded in such representations is already erased before second-phase pre-training. Therefore, it is also easier for GEEP to debias the model during further pre-training. We note that GEEP can lead to a slight increase of the original model's parameter size. We report the scale of the increase and its effect in Appendix~\ref{appx:capacity}.


\section{Experiments}
\vspace{-1mm}
In this section, we present the results of GEEP and its baselines to show that GEEP achieves state-of-the-art performances on gender fairness tasks while alleviating the forgetting issue of the baselines. 
\label{experiments}
\subsection{Experimental Setup}

In our experiments, we mainly use the publicly released RoBERTa-base model as the pre-trained model. We have also conducted experiments on publicly released BERT during preliminary explorations. Details on BERT experiments are in Appendix~\ref{appx:bert}.
Given a pre-trained RoBERTa-base model, we compare GEEP with two main baselines. The first baseline is the pre-trained RoBERTa-base model without any further training. 
The other important type of baselines are SPPA methods. For a fair comparison, our SPPA baseline uses the same gender-neutral data set that we construct for GEEP (details in Section 3.2) to further update all model parameters of the pre-trained RoBERTa-base. 
The detailed hyper-parameter settings of GEEP and SPPA can be found in Appendix~\ref{appx:hyper}.

\vspace{-2mm}
\subsection{Evaluation Tasks}
\vspace{-1mm}
To assess gender fairness, 
we conduct pronoun coreference resolution experiments on different data sets, Winogender \citep{rudinger-EtAl:2018:N18}, Winograd Schema Challenge (WSC) \citep{levesque-pkrr12}, and Definite Pronoun Resolution (DPR) \citep{rahman-emnlp12}. 
Pronoun coreference resolution is the task of linking the pronouns with their references in a text. 
\begin{table}
\caption{Results on Coreference Resolution task. }
\centering
\small\addtolength{\tabcolsep}{-2pt}
\begin{tabular}{lccc}  
\toprule
Data& RoBERTa &SPPA &GEEP \\ \hline
Winogender&50.9&57.3&\textbf{64.5}\\\hline
WSC&50.1&50.9&\textbf{52.7}\\\hline
DPR/WSCR&50.8&51.1&\textbf{53.6}\\
\bottomrule
\end{tabular}
\label{coref}
\vspace{-3mm}
\end{table}
In order to resolve a pronoun accurately, a model needs to overcome the biased link between gender and profession (e.g. the assumption that nurses are female) and instead make the decision based on available linguistic cues.
Therefore, better performances on pronoun coreference resolution usually indicates less gender bias preserved in the model \citep{kurita-gbnlp19}. Detailed setups of this experiment can be found in Appendix~\ref{appx:coreference}. Additionally, we also qualitatively and quantitatively evaluate our method on direct pronoun prediction. Details of this experiment are in Appendix~\ref{appx:pronoun}.
We note that given all existing tasks are designed for binary gender pronouns, we are unable to include all existing gender identities in our main experiments. We present an analysis on more gender identities in Appendix~\ref{appx:ac}. 


\begin{table}
\caption{GLUE results of different models.}

\centering
\small\addtolength{\tabcolsep}{2pt}
\begin{tabular}{l|c|cc}  
\toprule
Task &RoBERTa &SPPA& GEEP \\ \hline
MNLI&87.7&87.2&\textbf{87.7}\\\hline
QNLI&92.4&\textbf{92.4}&\textbf{92.4}\\\hline
QQP&91.8&91.3&\textbf{91.7}\\\hline
SST-2&95.4&94.7&\textbf{95.4}\\\hline
CoLA&64.1&38.9& \textbf{50.5}\\\hline
MRPC&91.4&88.8&\textbf{89.8}\\\hline
RTE&78.4&60.2&\textbf{68.7}\\\hline
STS-B&90.7&88.3&\textbf{89.9}\\\hline
AVG& 86.5& 80.2& \textbf{83.3}\\

\bottomrule
\end{tabular}
\label{gluetask}

\vspace{-5pt}
\end{table}

To evaluate how much each debiased model forgets after second-phase pre-training, 
we report the performances of the debiased models on GLUE benchmark \citep{wang-emnlp18}. Detailed setups of this experiment can be found in Appendix~\ref{appx:glue}.

\subsection{Results}
\vspace{-2mm}

We first show the pronoun coreference resolution results of different models on three datasets in Table 1.
Results show that GEEP model obtains the best accuracy compared to other models, especially on the Wingender dataset where the candidate nouns are professions. We also conduct an ablation study to show the effect of total training iterations on the performances of both methods. We find that GEEP can improve the model's performance with significantly fewer number of training iterations.
Details are in Appendix~\ref{appx:hyper}.


\



Then we show in Table \ref{gluetask} the performance of different models on $8$ GLUE tasks, to see how severe the forgetting issue is after the second-phase training of SPPA and GEEP. Compared with RoBERTa, SPPA suffers from forgetting issue in $7$ out of $8$ tasks except QNLI. 
For tasks like CoLA and RTE, the model's performance drops significantly (more than 10 points) after SPPA.
For tasks with larger data set for fine-tuning, such as MNLI, QQP and SST-2, they are less vulnerable to the quality of pre-training ~\citep{wu2020taking,joshi2020spanbert}. Therefore, SPPA's performance drop on such data sets is less significant.  
GEEP mitigates the forgetting issue of SPPA in all sub-tasks. Since GEEP ditches the original pre-trained profession embeddings and uses a smaller data set to update new profession embeddings, the forgetting problem cannot be fully avoided. While GEEP still achieves an average GLUE score of 83.3, significantly outperforming SPPA. 
We have also included an empirical analysis regarding to the reasons behind SPPA's GLUE performance drop in Appendix~\ref{appx:r1}.

\section{Closing Remarks}
\vspace{-2mm}
In this paper, we proposed GEEP to improve gender fairness of pre-trained language models with less catastrophic forgetting. 
For a fair comparison to existing work under the current gender fairness literature, we mainly conduct experiments with profession-related gender neutral data because profession-related gender bias is relatively more well studied so far. 
The good empirical results indicates it is worth to try applying GEEP to other more challenging and under-explored aspects of gender fairness, which would be our future work.
\bibliography{anthology}
\bibliographystyle{acl_natbib}
\clearpage
\appendix

\section{Limitations}
In this paper, we only focus on investigating and improving gender fairness of pre-trained language models and didn't touch other fairness issues given the length of the paper. 
However, we would like to note that with the investigation of other fairness issues in human language more deeply conducted, if the biased words regarding other fairness issues can be more specifically concluded, GEEP can be directly applied to address other fairness problems in pre-trained large language models.   
\clearpage

\section{Appendix}
\label{sec:appendix}
\subsection{Hyper-parameters for SPPA and GEEP}
\label{appx:hyper}
For the main results presented in the paper of second-phase pre-training in GEEP and SPPA, we further train RoBERTa-base
for $100,000$ steps with our gender-neutral data. We use an AdamW optimizer with a
learning rate of $1e-5$, max\_seq\_length of $128$ and batch size 256. In GEEP method, we initialize the embedding of every profession prompt with a normal distribution and standard deviations of $0.2$. 

Alongside the final results, we also evaluate SPPA and GEEP during the second-phase pre-training. In Table~\ref{coref-appx} we show SPPA and GEEP's performance on pronoun coreference resolution at the 20k iteration and 50k iteration. 
From Table~\ref{coref-appx} we can see that GEEP improves the pre-trained model's gender fairness with much less number of iterations. At 20k iteration, GEEP's performance is already better than SPPA's final performance on all 3 tasks. At 50k iteration, GEEP's performance has almost converged to its final scores on all 3 tasks. While SPPA's performance is still far behind its final performances on Winogender and WSC.
\subsection{Pronoun Coreference Resolution Experiment Setup}
\label{appx:coreference}
Pronoun Coreference Resolution is the task of linking the pronouns with their references in a text.  Studies show that BERT performance decreases in a text where the gender pronoun is female and the topic is biased towards the male gender \citep{kurita-gbnlp19}. To assess the performance of different models in pronoun coreference, we
fine-tune our models with GAP data set \citep{webster-acl18}.  We fine-tune each model for one epoch with a train batch size of $64$ and a learning rate of $5.0e-6$. After fine-tuning, we evaluate the performance of different models on three data sets:
\vspace{-3mm}
\begin{itemize}
    \item Winogender: This dataset includes $1,584$ sentences with three mentions: a profession, a participant, and a pronoun (where the pronoun is referred to either profession or pronoun)\citep{rudinger-EtAl:2018:N18}.
    \item WSC: The Winograd Schema Challenge (WSC) incorporates $273$ sentences used for commonsense reasoning for resolution
    \citep{levesque-pkrr12}.
    \item DPR: The Definite Pronoun Resolution (DPR) corpus with $131$ test sentences contains examples with two noun phrases and a pronoun or possessive adjective referring to one of the noun phrases \citep{rahman-emnlp12}.
\end{itemize}

\subsection{GLUE Experiment Setup}
\label{appx:glue}
To evaluate how much each debiased model forgets after second-phase pre-training, 
we fine-tune the pre-trained models on GLUE (\textbf{G}eneral \textbf{L}anguage \textbf{U}nderstanding \textbf{E}valuation)  to evaluate the performance of the pre-trained models. We follow previous work to use eight tasks in GLUE, including CoLA, RTE, MRPC, STS, SST, QNLI, QQP, and MNLI. For evaluation metrics, we report Matthews correlation for CoLA, Pearson correlation for STS-B, and accuracy for other tasks. We use the same optimizer (Adam) with the same hyper-parameters as in pre-training. Following previous work, we search the learning rates during the fine-tuning for each downstream task. For a fair comparison, we do not apply any published tricks for fine-tuning. Each configuration is run five times with different random seeds, and the \emph{average} of these five results on the validation set is calculated as the final performance of one configuration. We report the best number over all configurations for each task.


\begin{table*}
\caption{The average accuracy of different models on Coreference Resolution task. The best results are in bold.}
\centering
\small\addtolength{\tabcolsep}{-2pt}
\begin{tabular}{l|c|cc|cc|cc}  
\toprule
Data& RoBERTa &SPPA-20k &GEEP-20k & SPPA-50k & GEEP-50k & SPPA-100k & GEEP-100k \\ \hline
Winogender&50.9&51.6&64.3&54.6&64.5&57.3&\textbf{64.5}\\\hline
WSC&50.1&50.1&52.1&50.5&52.3&50.9&\textbf{52.7}\\\hline
DPR/WSCR&50.8& 50.9&52.1&51.1&53.4&51.1&\textbf{53.6}\\\hline
Avg GLUE & 86.5 & 82.7 & 85.9 &80.7 &84.5& 80.2&\textbf{83.3}\\
\bottomrule
\end{tabular}
\label{coref-appx}
\vspace{-12pt}
\end{table*}


\subsection{Pronoun Prediction Experiment Setup and Results}
\label{appx:pronoun}
\begin{figure*}
    \centering
        \includegraphics[width=0.8\textwidth]{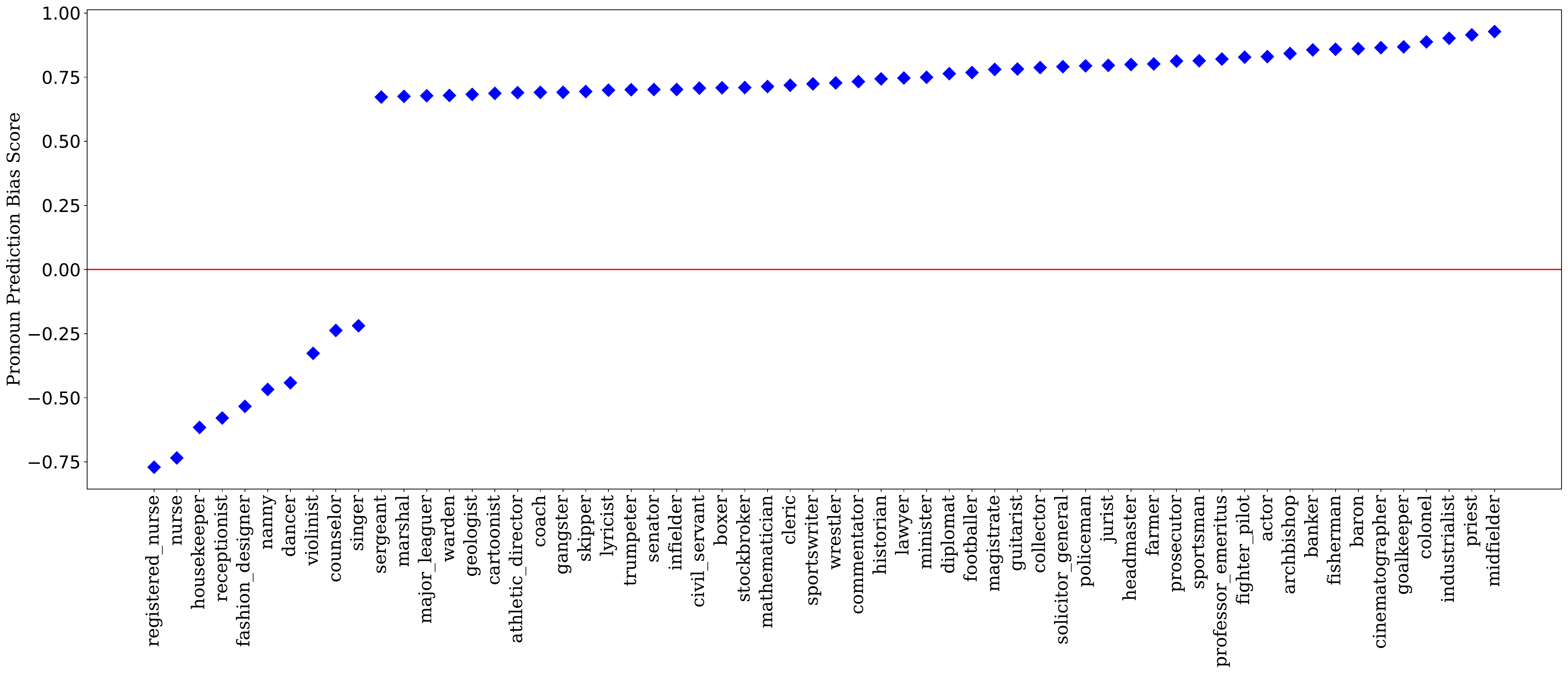}
        \vspace{-3mm}
        \caption{An example of gender bias in $60$ most biased profession words in BERT-base model. For each profession, we measure the difference between the probability of filling the masked pronoun in each template sentence with "he" and "she" tokens.
        Some words such as nurse (-0.73) and receptionist (-0.57) are supposed to be gender neutral by definition but BERT-base model consider them as female professions. On the other hand, lawyer (0.74) and prosecutor (0.81) are considered as jobs for males.}
    \label{fig:bert-base}
\end{figure*}

Different approaches have been proposed to
quantify and analyze the gender bias in contextual language models
\citep{de-acl21,webster-arxiv20,kurita-gbnlp19}.
For BERT, we choose one approach that can be directly applied to a model pre-trained with Masked Language Modeling (MLM) loss without further fine-tuning. In this approach, we first define a template containing a pronoun and a profession. The profession is supposed to be gender-neutral however it is currently viewed with gender bias to a large extent. By masking the pronoun, the model is queried to predict the pronouns at the masked position given the context, including the profession. Here is an example, ``[MASK]'' is a registered nurse. The difference between the probabilities of filling the masked position in each sentence with "he" and "she", is used to show gender bias in the model,
\begin{align}
    \text{Pronoun}&\text{ Bias Score}= \\ &\text{Prob}(\textbf{"he"})- \text{Prob}(\textbf{"she"}).
\end{align}
To assess fairness in BERT model, we consider $303$ of professions used by \cite{bolukbasi-neurips16}. 
In our study,
we analyze a public available pre-trained BERT-Base model \footnote{https://github.com/google-research/bert} that contains 12 layers, 768 hidden nodes, 12 heads, and 110M parameters.
Figure \ref{fig:bert-base} shows gender bias of $60$ of such professions in BERT-base model.
Positive values mean that the professions are biased towards male and vice versa. As the plots show, the contextual representations of professions in BERT-base model exhibits strong gender bias. Professions such as nurse and housekeeper are viewed as jobs for females while 
surgeon and mathematicians are assumed to be jobs for males. 

\begin{figure*}
\vspace{-50pt}
\centering
\begin{subfigure}{\textwidth}
    \centering
    \includegraphics[height=1.8in]{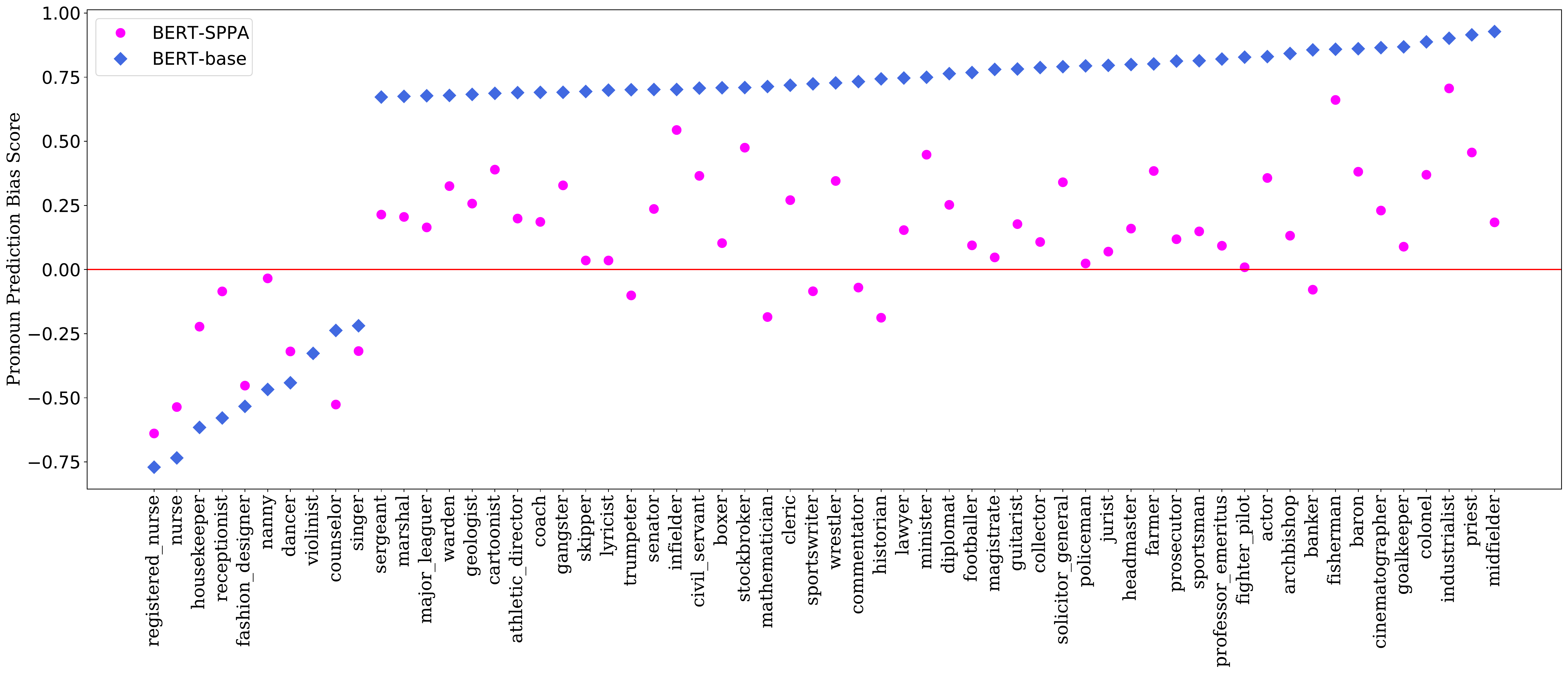}
    \caption{ Comparison between pronoun prediction bias in SPPA and BERT-base models
    }\label{fig:image1}
\end{subfigure}
    \hfill
\bigskip
\begin{subfigure}{\textwidth}
  \centering
  \includegraphics[height=1.8in]{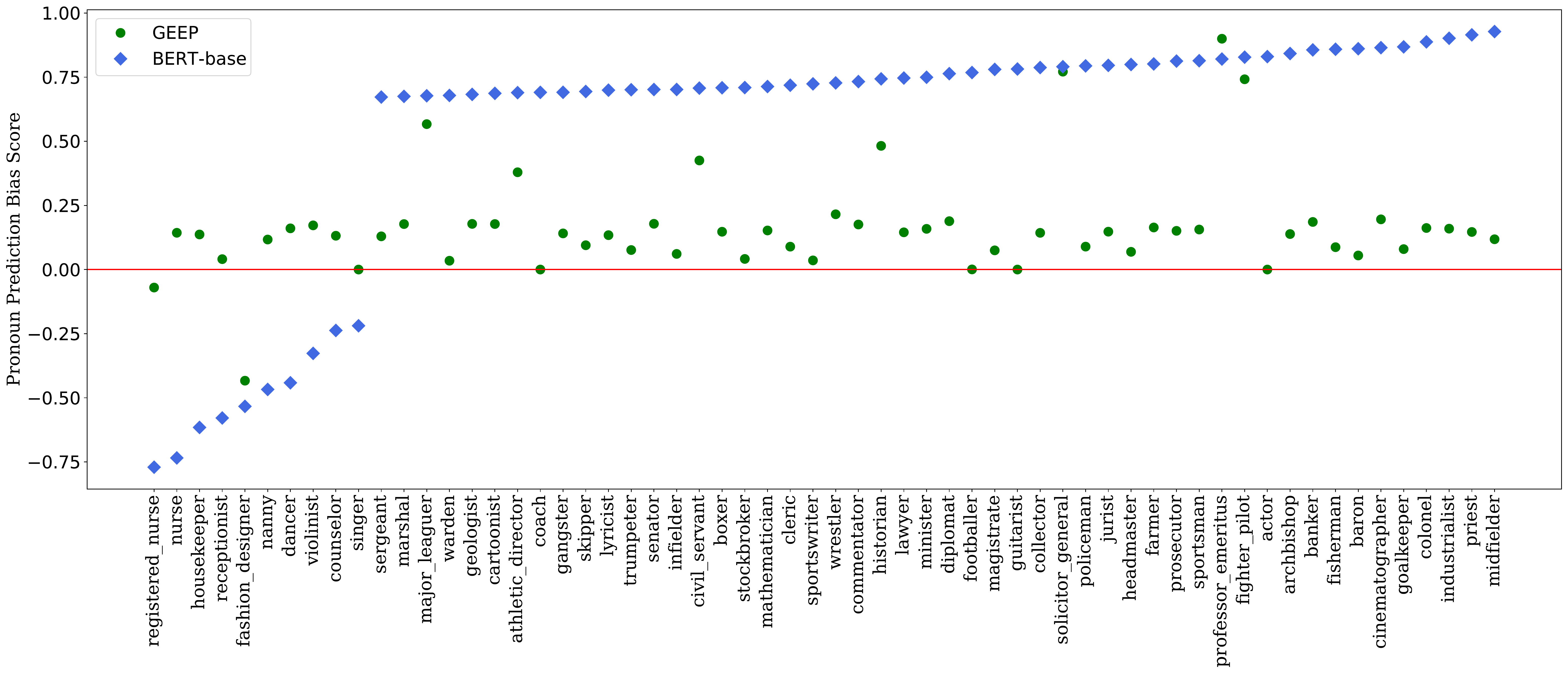}
  \caption{Comparison between pronoun prediction bias in GEEP and BERT-base models}\label{fig:image3}
\end{subfigure} 
\caption{Difference between the probabilities of filling a masked pronoun with "he" and "she" tokens in the template sentences containing 60 most biased professions. GEEP method outperforms the two other methods. For example,
the bias score for "nurse" token decreases from $-0.7$ in BERT-base to $-0.5$ in BERT-SPPA and $0.1$ in GEEP model.}
\label{fig:mask_filling}
\end{figure*}

To find the reference of each pronoun in the template sentences, we follow \cite{kocijan-acl19} approach.
Specifically, during the evaluation for every data set, in each sentence there are two candidate nouns (such as ``nurse'' or ``surgeon'') and a pronoun. The pronoun is replaced with a [MASK] token, and the model makes a prediction at the masked pronoun position from the two candidate nouns. In order to resolve a pronoun accurately, a model needs to overcome the biased link between gender and profession (e.g. a normative assumption that nurses are female) and instead make the decision based on the available linguistic cues. We report the prediction accuracy of all $3$ methods on the aforementioned three data sets. 

Figure \ref{fig:mask_filling} displays the pronoun prediction bias score (defined in Equation 5) of all methods for 60 biased professions defined in \citep{bolukbasi-neurips16}. Specifically, in both sub-figures, blue dots show the pronoun prediction bias score from BERT-base model for each profession. In Figure \ref{fig:mask_filling} (a), the pink dots are the bias scores from BERT-SPPA model. We can see from this sub-figure that compared with BERT-base, the bias scores from BERT-SPPA model are indeed closer to $0$, indicating that BERT-SPPA can mitigate gender bias of such professions to some extent. In Figure \ref{fig:mask_filling} (b), the blue dots are the bias scores from GEEP model. Compared with both BERT-SPPA and BERT-base, GEEP's bias scores are significantly closer to $0$, indicating that GEEP is more effective at removing gender bias from such biased professions compared with BERT-SPPA. 
Moreover, we also calculate the
average absolute pronoun prediction bias score for all $303$ gender-neutral profession words in \citep{bolukbasi-neurips16}. We obtain 0.44 for BERT-base, 0.16 for BERT-SPPA and 0.13 for GEEP. GEEP model gets the lowest average bias with $70\%$ reduction compared to the BERT-base model.

\subsection{Analysis regarding SPPA's performance drop on GLUR}
\label{appx:r1}

We conduct experiments to analyze reasons behind the GLUE performance drop of SPPA demonstrated in Table 2 in our original submission. The performance drop of SPPA compared to RoBERTa can be of two reasons: 1) the model is further trained with a subset of Wikipedia significantly smaller than the RoBERTa pre-train data, which could enforce the model to forget about the information embedded in the large RoBERTa pre-train data; 2) we processed the subset of Wikipedia to make them gender-neutral, which could potentially introduce noise and distribution mismatch with the downstream data. 
To provide a more detailed analysis, we conduct experiments as follows. 

First, starting from a pre-trained RoBERTa, we further train the model with SPPA on the same subset of Wikipedia that we used in main experiments without making the data subset gender-neutral. 
We name this model SPPA-without-GN (Gender Neutralization). We also run GEEP-without-GN to see whether GEEP can still alleviate forgetting when the data is just small but not debiased. For GEEP-without-GN, we further train a RoBERTa with the same Wiki subset without gender neutralization. During this further training of GEEP-without-GN, we follow GEEP to add and update new profession embeddings while freezing the rest entire model. GLUE results of SPPA-without-GN and GEEP-without-GN are in Table~\ref{rebuttal} in this pdf.

By comparing SPPA, SPPA-without-GN and the original RoBERTa, we can find SPPA-without-GN performs better than SPPA while  worse than RoBERTa. It suggests that both data subset selection and gender neutralization contribute to the performance drop of SPPA compared to RoBERTa. 
We would also like to note that GEEP-without-GN outperforms SPPA-without-GN as well and achieve similar GLUE score as RoBERTa.  This indicates that GEEP can also alleviate forgetting introduced by data subset selection effectively when there is not gender-neutralizing procedure is taken. 

\begin{table*}
\caption{GLUE results. The best results among SPPA and GEEP are in bold.  }
\centering
\small\addtolength{\tabcolsep}{2pt}
\begin{tabular}{l|c|cc|cc|c}  
\toprule
Task &RoBERTa &SPPA& GEEP &SPPA-without-GN & GEEP-without-GN & SPPA-with-NPE\\ \hline
MNLI&87.7&87.2&\textbf{87.7}&87.3&87.7 &87.2\\\hline
QNLI&92.4&\textbf{92.4}&\textbf{92.4}&92.3&92.4 &92.3\\\hline
QQP&91.8&91.3&\textbf{91.7}&91.4&91.8&91.5\\\hline
SST-2&95.4&94.7&\textbf{95.4}&95.0&95.4&94.7\\\hline
CoLA&64.1&38.9& \textbf{50.5}&40.2&59.6&39.3\\\hline
MRPC&91.4&88.8&\textbf{89.8}&88.8&90.5&88.8\\\hline
RTE&78.4&60.2&\textbf{68.7}&66.4&73.1&61.0\\\hline
STS-B&90.7&88.3&\textbf{89.9}&89.5&90.4&88.5\\\hline
AVG& 86.5& 80.2& \textbf{83.3}&81.4&85.1&80.4\\
\bottomrule
\vspace{-20pt}
\end{tabular}
\label{rebuttal}
\end{table*}

\subsection{Discussions on non-binary gender identities}
\label{appx:ac}

In this discussion, we would like to start with the pronoun choices for different gender identities. Because in our submission we mainly try to address the unfair pronoun preference of pre-trained models.
According to social research,
gender-neutral pronouns are more appropriate for referring to transgender and non-binary individuals \cite{deutsch2015electronic}. ‘Zie’ and ‘hir’ are specific to transgender community, but people outside of the community are not familiar with these pronouns. 
\citet{deutsch2015electronic} has proposed a Gender-ID to pronoun mapping for transgenders and Genderqueer in electronic health records (EHR). In this system, transgenders are mapped to he/his or she/her where there exists gender bias, but genderqueer are mapped to they/them.  
For people who prefer binary pronouns(he/she) regardless of their gender identities, our experiments still hold because the pronoun
coreference resolution tasks that we evaluate on, i.e. Winogender, WSC and DPR/WSCR, are all binary-pronoun tasks.

However, an alternative to asking for preferred pronouns would be to use singular pronouns to address everyone until the individual indicates a preference to use certain pronouns and/ or reveal their gender identity \cite{darr2016pronouns}.
One optional term that is already used as a singular pronoun like "they/their" \cite{darr2016pronouns,richards2016non,sun2021they}. If such singular pronoun can be promoted to a larger community, the pronoun unfairness issue can be resolved from the data fundamentally.

\subsection{The capacity increase of GEEP compared to SPPA}
\label{appx:capacity}

By adding profession embeddings, it is true that the total number of model parameters slightly increases. However, the entire size of the newly-added parameters is 303*768=232k, which is only 0.21\% of the original RoBERTa parameter size (110 million). 303 is the number of professions and 768 is the embedding size of RobERTa. Therefore, even if we extend this method to other fairness problems in the future and add more new word embeddings such as $3000$ words or $10000$ words, the newly-added parameters would be just around 2\% or 9\% of the original parameter size, which wouldn't cause serious scaling issue.  

Moreover, we run a new SPPA variant that has the same capacity (the same number of parameters) with GEEP. In the new SPPA variant, we conduct SPPA training after adding new word embedding of the profession names, same as GEEP. We refer this model  as SPPA-with-NPE (new profession embeddings). The difference between SPPA-with-NPE and GEEP is GEEP's core implementation to prevent forgetting, that GEEP freezes the rest parameters during further training and only update new profession embeddings. While SPPA-with-NPR updates all parameters including the original model parameters and the newly added profession embeddings. 
When encountering the pre-defined profession names in training or fine-tuning, SPPA-with-NPR also updates their new embeddings instead of old word/token embeddings.
GLUE results are shown in Table~\ref{rebuttal}. Compared with SPPA, SPPA-with-NPE can alleviate forgetting slightly and achieve better debiasing results, while still significantly under-perform GEEP. Results on pronoun coreference resolution tasks show the same trend. SPPA-with-NPE got 58.6 on Winogender, 51.3 on WSC and 52.4 on DPR/WSCR. They are all slightly better than SPPA while significantly lower than GEEP.

\subsection{Quality of gender-neutral data}
\label{appx:grammar}

The relatively big performance drop of both our method and SPPA compared to the original RoBERTa motivates us to analyze more on the quality of our gender-neutral data.

While first we note that CoLA and RTE are known to be more sensitive to quality of pre-trained models compared with other tasks in GLUE, due to their small data sizes. In other words, if the pre-trained model is trained insufficiently or with less data, we can see a larger performance drop on CoLA and RTE compared with other tasks. While if the pre-trained model's quality is better, we can see larger improvements on them as well. This trend has been observed in  BERT vs  RoBERTa, BERT vs Span-BERT, and BERT vs ELECTRA. 
Therefore, the reason for the large performance drop on COLA can partially be its natural sensitivity to our small data size of further training RoBERTa.

Second, the gender neutralization process of the training data could cause gender mismatch between pronouns and some very rare nouns.
we did follow the reviewer's suggestion to sample 500 sentences from the augmented dataset and manually checked whether there are grammar errors. In these 500 sentences, there are no grammar errors, such as mismatches between nouns and verb formats (e.g. "he are"). Because during the gender neutralization, we follow previous work to just swap the gender-related pronouns (such as he/she) or nouns (such as uncle/aunt) when profession names occur. And such gender-related nouns share the same verb formats with their counterparts. We also share the full list of gender-related nouns in the appendix in this submission.
However, when we sample more modified sentences, we find that if a rare gender-related noun, such as ``spinster'', that is not on the published gender-related noun list occurs, the gender neutralization process would change the pronoun while leave the noun unchanged since it is not on the list. 
Although it happens quite rarely, this causes pronoun misuse that could lead to grammar errors in pre-training data that contribute to the performance drop on CoLA.

\subsection{Experiment Results on BERT}
\label{appx:bert}
\begin{table}
\caption{GLUE results.  The best results are in bold.  }
\centering
\small\addtolength{\tabcolsep}{2pt}
\begin{tabular}{l|c|c c}  
\toprule
Task& BERT-base &BERT-SPPA& GEEP \\ \hline
MNLI&84.3&84.0&\textbf{84.1}\\\hline
QNLI&91.4&90.0&\textbf{91.3}\\\hline
QQP&90&90.1&\textbf{90.4}\\\hline
SST-2&93&92.2&\textbf{92.4}\\\hline
CoLA&54.0&52.0&\textbf{53.0}\\\hline
MRPC&85.7&84.1&\textbf{84.9}\\\hline
RTE&69.4&\textbf{69.8}&69.1\\\hline
STS-B&88.0&\textbf{88.0}&87.0\\\hline
AVG &82.0 & 81.3&\textbf{81.6}\\
\bottomrule
\end{tabular}
\label{gluetask}

\end{table}
\begin{table}
\caption{The average accuracy of different models on Coreference Resolution task. The best results are in bold.}

\small\addtolength{\tabcolsep}{-2pt}
\begin{tabular}{lccc}  
\toprule
Data& BERT-base &BERT-SPPA &GEEP \\ \hline
Winogender&50&50.7&\textbf{62.9}\\\hline
WSC&50.1&50.2&\textbf{50.5}\\\hline
DPR/WSCR&50.7&50.9&\textbf{52.8}\\
\bottomrule
\end{tabular}
\label{coref}
\vspace{-10pt}
\end{table}
During the preliminary exploration on this problem, we have also applied SPPA and GEEP on publicly released BERT and conducted pronoun coreference resolution and GLUE experiments on them. In this experiment, we only further trained the released BERT model for 10k iterations with our gender-neutral data.
Moreover, our gender-neutral data set (7.1 GB) is not significantly smaller than the original pre-training data of BERT (16 GB), and  the two data sets both come from Wikipedia. Due to these two reasons, the forgetting problem on this BERT experiment is not as obvious for SPPA.

Table \ref{gluetask} shows the performance of different methods on $8$ GLUE tasks. 
Although the forgetting is less server, SPPA still suffers from forgetting issue in the following $6$ tasks out of the total $8$ tasks, CoLA, MRPC, STS-B, MNLI, QNLI, and SST-2. As for the average GLUE score, SPPA is $0.7$ point lower after its second-phase pre-training, which is not a small margin considering it is the average score of $8$ tasks. GEEP mitigates the forgetting issue of SPPA in all sub-tasks except in RTE. GEEP also gets the average GLUE score of $82.8$, which outperforms SPPA and is similar to the original GLUE score of the pre-trained BERT.

Table \ref{coref} shows the coreference resolution results of different models on three data sets.
Results show that GEEP model obtains the best accuracy compared to other models, especially in Wingender dataset where the candidate nouns are professions. We observe that the SPPA method also can help improve coreference resolution performance of the pre-trained model, but not as effective as GEEP.

\end{document}